\documentclass{article}

\usepackage{PRIMEarxiv}

\usepackage[utf8]{inputenc} 
\usepackage[T1]{fontenc}    
\usepackage{amsmath}
\usepackage{hyperref}       
\usepackage{cleveref}
\usepackage{url}            
\usepackage{booktabs}       
\usepackage{amsfonts}       
\usepackage{nicefrac}       
\usepackage{microtype}      
\usepackage{lipsum}
\usepackage{array} 
\usepackage{fancyhdr}       
\usepackage{graphicx}       
\graphicspath{{media/}}     
\usepackage{soul}
\usepackage{soul,xcolor}
\sethlcolor{yellow}
\usepackage{subcaption}
\usepackage{makecell}
\usepackage[square,authoryear]{natbib}

\hypersetup{
    colorlinks=true,
    citecolor=blue,
    linkcolor=blue,
    urlcolor=blue
}

\crefformat{table}{#2Table~#1#3}
\Crefformat{table}{#2Table~#1#3}
\crefformat{figure}{#2Figure~#1#3}
\Crefformat{figure}{#2Figure~#1#3}
\crefformat{equation}{#2Equation~#1#3}
\Crefformat{equation}{#2Equation~#1#3}
\crefformat{appendix}{#2Appendix~#1#3}
\Crefformat{appendix}{#2Appendix~#1#3}

\title{Skewed score:\\a statistical framework to assess autograders}

\author{
  \textbf{Magda Dubois\textsuperscript{1}}
  \quad \textbf{Harry Coppock} 
  \quad \textbf{Mario Giulianelli}
  \quad \textbf{Timo Flesch} \\ \\
  \quad \textbf{Lennart Luettgau\textsuperscript{2}} 
  \quad \textbf{Cozmin Ududec\textsuperscript{2}} \\ \\
  UK AI Security Institute \\ \\
  London, United Kingdom \\ \\
  \textcolor{gray}{\href{https://doi.org/10.48550/arXiv.2507.03772}{\textcolor{gray}{https://doi.org/10.48550/arXiv.2507.03772}} (arXiv:2507.03772)}
}

\begin{document}
\maketitle
\renewcommand{\thefootnote}{\arabic{footnote}} 
\footnotetext[1]{Corresponding author: magda.dubois@dsit.gov.uk}
\footnotetext[2]{These authors contributed equally as last authors}

\begin{abstract}
The evaluation of large language model (LLM) outputs is increasingly performed by other LLMs, a setup commonly known as ``LLM-as-a-judge'', or autograders. While autograders offer a scalable alternative to human evaluation, they have shown mixed reliability and may exhibit systematic biases, depending on response type, scoring methodology, domain specificity, or other factors. Here we propose a statistical framework based on Bayesian generalised linear models (GLMs) that enables researchers to simultaneously assess their autograders while addressing their primary research questions (e.g., LLM evaluation). Our approach models evaluation outcomes (e.g., scores or pairwise preferences) as a function of properties of the grader (e.g., human vs.\ autograder) and the evaluated item (e.g., response length or the LLM that generated it), allowing for explicit quantification of scoring differences and potential biases within a unified framework. In addition, our method can be used to augment traditional metrics such as inter-rater agreement, by providing uncertainty estimates and clarifying sources of disagreement. Overall, this approach contributes to more robust and interpretable use of autograders in LLM evaluation, enabling both performance analysis and bias detection.
\end{abstract}

\section*{Introduction}
\label{sec:intro}
 
Large language models (LLMs) are being deployed across a wide range of tasks (e.g., code generation, question answering, decision support, autonomous tool use like calling APIs to retrieve documents). To understand their capabilities and ensure reliable performance and safety, LLM outputs need to be continuously and systematically evaluated. Given the high volume and increasing complexity of these evaluations, automated approaches have become common. Unsurprisingly, LLMs themselves have emerged as candidates for this task, a practice referred to as ``LLM-as-a-judge'' \citep{zheng2023judging,liu-etal-2023-geval,chiang-lee-2023-large,fu-etal-2024-gptscore,bavaresco-2025-instead,wang2025can} or as autograding \citep{schneider2023towards,guan2024deliberative}. For brevity, we here refer to these models as autograders.\looseness-1

To assess the reliability of autograders, researchers typically compare their outputs to human judgments using metrics such as correlation coefficients or inter-rater agreement scores \citep{bavaresco-2025-instead,wang2025can}. These metrics provide useful summaries and show that autograders can disagree with human graders and with each other. An important open question is whether such disagreement reflects random noise or consistent patterns in autograder behaviour.\looseness-1

Recent studies suggest that autograders may indeed exhibit systematic biases. One such bias is self-bias, where LLM-based graders tend to assign higher scores to responses generated by the same LLM \citep{panickssery-2024-recognize,liu-etal-2024-llms-narcissistic}, or more broadly to machine-generated content over human-written responses \citep{liu-etal-2023-geval}. Another common issue is length bias, where longer answers are preferred regardless of their actual quality \citep{zheng2023judging,dubois2024lengthcontrolled}. Additional biases include preferences for certain writing styles, answer structures, or the presence of certain keywords \citep[e.g.,][]{koo-etal-2024-biases,wang-etal-2024-fair,stureborg-2024-inconsistent,wu-aji-2025-style}. Moreover, graders may exhibit intransitive preferences (e.g., preferring A over B, B over C, yet C over A), revealing inconsistencies that cannot be captured by standard grader evaluation metrics \citep{xu2025investigating}.

Existing methods for autograder evaluation are limited in their ability to isolate and quantify such biases and inconsistencies. They do not explain why disagreement arises, do not account for how grader behaviour may vary with grader identity, model characteristics, or context, and do not quantify uncertainty in these effects. While recent work has started employing statistical modeling (e.g., logistic regression for length bias \citep{dubois2024lengthcontrolled}), these approaches typically remain narrowly tailored to specific biases and do not offer an integrated evaluation framework. In this paper, we introduce a Bayesian generalised linear modeling framework that complements existing metrics and provides an interpretable, extensible, and statistically grounded analysis of grader behaviour.

In the following sections, we begin by presenting the general framework. We then walk through a series of common questions to demonstrate how the framework can be used to answer them. We begin by explaining how autograder scores can be compared to human scores (\hyperref[sec:q11]{Question 1.1}), and how this comparison can be integrated into an LLM evaluation analysis (\hyperref[sec:q12]{Question 1.2}). We then gradually increase the complexity of the setup and show how different questions about autograder quality can be addressed within this integrated framework. Specifically, we show how to assess whether the graders are biased towards certain models being evaluated (\hyperref[sec:q2]{Question 2}) and how to quantify individual differences in the case of multiple graders (\hyperref[sec:q3]{Question 3}). We also show how to analyse item-level patterns (\hyperref[sec:q4]{Question 4}), which allows us to compute inter-rater agreement metrics with quantification of uncertainty and, crucially, to identify whether disagreement stems from noise or systematic bias (something that traditional metrics cannot disentangle). Finally, we show how this framework can be applied to pairwise judgments settings, in which LLM outputs are compared directly to one another (as opposed to being assigned absolute scores), how to quantify intransitive (e.g., cyclic) preferences and how to assess whether graders have biases towards longer formats (\hyperref[sec:q5]{Question 5}). 
In \cref{tab:glm_questions}, we provide a summary of the evaluation questions along with their corresponding formalisations as a practical guide for researchers using our framework.\looseness-1

\section*{Proposed framework}

We propose using generalised linear models (GLMs) as a practical and flexible approach for analysing autograder performance during LLM evaluations. GLMs extend standard linear regression by allowing the outcome variable to follow a wider range of distributions, making them suitable for many data types beyond normally distributed outcomes. At their core, GLMs relate the expected outcome to a linear predictor via a link function:

\begin{equation}
\label{eq:lin}
g(\mu) = \beta_0 + \beta_1 X_1 + \beta_2 X_2 + \dots + \beta_n X_n
\end{equation}

where $\mu = \text{E}[Y]$ represents the expected outcome, $g(\cdot)$ is a link function appropriate to the distribution of the outcome, $\beta_0$ is the intercept, and each $\beta_i$ represents the effect of predictor variable $X_i$. This structure is particularly useful as it allows researchers to include multiple predictors, control for potential confounders, and isolate the contribution of each variable to the outcome.

In the context of LLM evaluations, the outcome might take different forms depending on the evaluation design. For example, if the outcome is an ordinal score (e.g., 1–10), it can be modelled using an ordered logistic likelihood, or if it is a binary preference between two model outputs, a binomial likelihood is more appropriate. In both cases, the outcome is modelled as a function of variables related to the evaluated item (e.g., which LLM produced the response, how long it was, etc.) and the grader (e.g., human vs.\ autograder). Including both types of predictors allows the model to control for confounding factors and quantify systematic differences in scoring behaviour.

Using a Bayesian approach provides full posterior distributions over model parameters rather than point estimates, enabling direct uncertainty quantification and more robust inference, especially in settings with limited data, noisy measurements, or complex dependencies (for a foundational reference on Bayesian GLMs, see \citet{mcelreath2018statistical}). This is particularly important in limited-data settings where confidence intervals based on the central limit theorem often underestimate uncertainty \citep{bowyer2025positiondontuseclt}. 

Moreover, LLM evaluation data often exhibit structured dependencies such as multiple annotations from the same (or similar) graders, or items targeting related domains (e.g., several questions assessing historical knowledge). GLMs naturally allow for hierarchical extensions that can account for these dependencies, enabling more robust inference and better-calibrated uncertainty \citep{luettgau2025hibayeshierarchicalbayesianmodeling}.

Importantly, a GLM-based framework supports answering the primary evaluation question (e.g., ``how well does my LLM perform?'') while simultaneously identifying potential biases introduced by autograders. In the following sections we present illustrative examples of how these methods can be applied in practice.

\begin{table}[h!]
\centering
\renewcommand{\arraystretch}{1.3}
\begin{tabular}{
    >{\raggedright\arraybackslash}p{5.2cm}|
    >{\raggedright\arraybackslash}p{6.5cm}|
    >{\centering\arraybackslash}p{2.2cm}
}
\toprule
\textbf{Evaluation question} & \textbf{GLM implementation} & \textbf{Explained in} \\
\midrule
What grades does my autograder give compared to a human grader? & Include grader as a main effect. & \hyperref[sec:q11]{Question 1.1}\\
Can I assess the quality of my autograder during my evaluations? & Include both grader and LLM as predictors. & \hyperref[sec:q12]{Question 1.2}\\
Do my autograder(s) favour their own output? & Include an interaction between grader and LLM; test for model-aligned preferences. & \hyperref[sec:q2]{Question 2}\\
Is there a general human vs autograder difference? & Use a hierarchical GLM with grader-level effects nested in grader type (e.g., human vs.\ autograder). & \hyperref[sec:q3]{Question 3}\\
Are some graders more lenient or strict than others? & Estimate individual grader effects; inspect variation across graders. & \hyperref[sec:q3]{Question 3}\\
Do some items consistently receive higher or lower scores? & Include item as a predictor; test whether some items are systematically easier or harder. & \hyperref[sec:q4]{Question 4}\\
Do graders disagree more on some items than others? & Include grader x item interaction; test for grader-specific scoring patterns across questions. & \hyperref[sec:q4]{Question 4}\\
What is the uncertainty around inter-rater agreement metrics? & Simulate scores from the model and compute agreement (e.g., Krippendorff’s $\alpha$) with uncertainty. & \hyperref[sec:q4]{Question 4}\\
Do grader(s) favour longer responses? & Include token length (or token length difference) as a predictor. & \hyperref[sec:q5]{Question 5}\\
Do my grader(s) exhibit intransitive (e.g., cyclic) preferences? & Estimate pairwise probabilities and compare them across pairs. & \hyperref[sec:q5]{Question 5}\\
Is my grading scale well calibrated? & Inspect cutpoints from ordered regression to analyse spacing and interpretability of score intervals. & \cref{sec:grading_scale}\\
\bottomrule
\end{tabular}
\vspace{0.7em}
\caption{Overview of evaluation questions, their GLM-based implementation, and corresponding paper sections.}
\label{tab:glm_questions}
\end{table}

\section*{Illustrative examples}
Imagine a typical scenario: a researcher, Florence\footnote{In tribute to the pioneering work of two Florence Nightingales in statistics: the 19th-century nurse who applied statistical methods to public health and the 20th-century statistician who advanced combinatorics and statistical theory.}, is using LLMs to answer open-ended questions.
To evaluate the quality of the model's responses, she decides to manually grade them. She develops a detailed rubric (\cref{tab:rubric}) with scores ranging from 1 (completely off-topic) to 10 (demonstrating deep understanding and insightful connections).

Florence carefully evaluates N=100 model responses using her rubric. For example, consider the question: ``During World War II, which part of aircraft did statisticians decide to reinforce?''. A low-scoring response (e.g., 3 points) might simply say: ``The areas with the least bullet holes.'' While this demonstrates some understanding, it misses the important statistical reasoning behind this strategy. In contrast, a high-scoring response (e.g., 9 or 10 points) would explain that they focused on areas with no damage, because aircraft hit in those places often did not survive to return. The response might further elaborate on the statistical principle of survivorship bias \citep{wald1939contributions} and its practical implications. 

This manual evaluation process proves to be very time-consuming, so Florence decides to turn to an autograder. She provides it with her rubric and a few examples of graded responses. By outsourcing the grading of open-ended answers to an autograder, she makes the assessment of the LLM's question-answering capabilities not only more efficient but also significantly more scalable. \looseness-1

This could have been the end of it, but being a careful researcher, she naturally asks herself, ``Can I actually trust my autograder?''. This central question unfolds into several more specific ones, for example: ``How do the autograder's scores compare to mine?'', ``Does the autograder favour generations produced by LLMs from its own model family?'', ``Does it tend to reward longer answers?''. Using a GLM framework, she can formalise and answer these questions while simultaneously evaluating the LLM's performance in answering open-ended questions. \looseness-1

To do this, Florence needs to build a set of statistical models, select the one that best explains her data, and examine how different covariates (e.g., grader, LLM, prompt, response length) and their interactions among one another systematically influence the scores. For instance, by including a human (vs.\ non-human) variable, she might discover that the autograder gives scores systematically 2 points higher than she does. Applying this approach across different covariates and interactions allows her to identify specific biases (such as model favouritism or length preferences), understand how grading patterns arise, and make informed adjustments to the autograder or evaluation process. 

To illustrate how the framework can be applied in practice, we will follow Florence as she navigates common challenges in evaluating LLM outputs with autograders. All examples are based on simulated data, which can be reproduced from a publicly available  \href{https://github.com/magda-dubois/skewed-score}{repository}. To facilitate wider adoption, all GLMs presented in this paper are implemented in the open-source \href{https://github.com/UKGovernmentBEIS/hibayes}{HiBayes package}.  \cref{tab:glm_questions} summarises the evaluation questions addressed, their corresponding GLM formulation, and the sections where each is presented and explained in detail.

\section*{Question 1: How do scores from an autograder compare to scores from a human expert?}
\phantomsection
\label{sec:q1}

\subsection*{1.1 Quantify the mean difference} 
\begin{figure}
  \centering
  \includegraphics[width=0.9\textwidth]{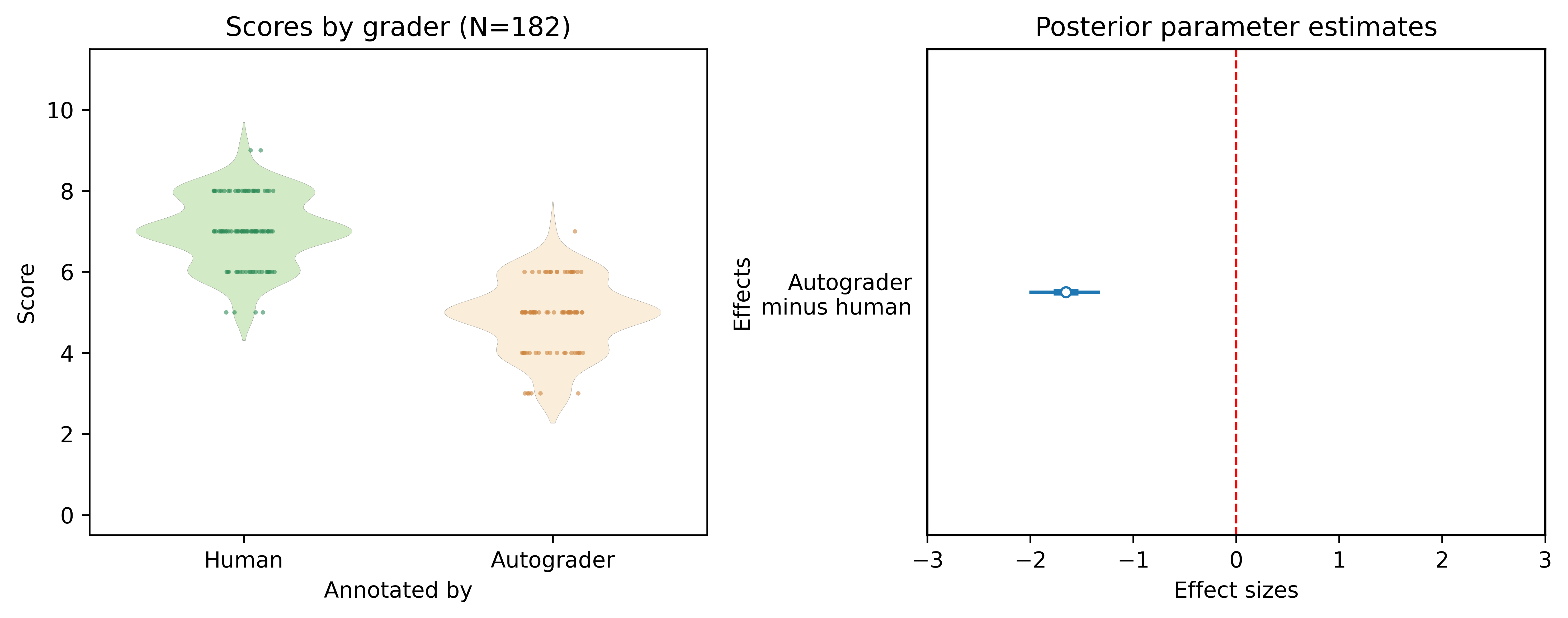}
  \caption{Illustration of how to use Bayesian GLMs to address \hyperref[sec:q11]{Question 1.1} (quantifying the mean difference between scores assigned by an autograder and a human grader) using simulated data. Left panel: Violin plot of simulated scores for LLM-generated answers graded by a human expert (Florence in the example) and an autograder. Right panel: Posterior distributions of estimated effects. The horizontal blue lines represent 95\% credible intervals. The dashed red vertical line indicates a null effect ($\beta = 0$). The coefficient for autograder minus human is negative, and its credible interval does not include zero, indicating strong evidence that autograders assign lower scores than human graders.}
  \label{fig:fig_autograder}
\end{figure}

\phantomsection
\label{sec:q11}
As Florence wants to validate the autograder before using it, she decides to run it on the same N=100 examples that she had graded, and compare the results. In the left panel of \cref{fig:fig_autograder}, we see a violin plot of the simulated scores given by Florence (human expert) in green, and by the autograder in yellow. Using our GLM framework we can fit a regression model with parameter estimates that quantifies the effect of the grader on the assigned score (\cref{eq:model_autograder}). Because the score can take values [1-10], we use an ordered logistic likelihood function, which accounts for the finite range and the categorical nature of the outcomes. This ordered logistic model has two main components: the linear predictor (cf. \cref{eq:lin}) and the cutpoints. In our case, the linear predictor produces a continuous value on a latent (unobserved) scale. The ordered logistic model assumes that our discrete 1-10 scores are actually observations of this underlying continuous scale, divided by cutpoints. When the linear predictor falls between particular cutpoints $c_j$, we observe the corresponding discrete score. For example, if the linear predictor value falls between cutpoints $c_3$ and $c_4$, we would observe a score of 4. These cutpoints are estimated from the data along with the $\beta$ coefficients (see \cref{sec:grading_scale} for more technical details).

In the following sections, we will focus primarily on the linear predictor component of the model, as the $\beta$ coefficients directly capture the effects of the variables of interest on the latent scale.

\begin{equation}
\begin{gathered}
\text{score}_i \sim \text{OrderedLogistic}(\phi_i, \boldsymbol{c}) \\
\phi_i = \beta_0 + \beta_1 \cdot X_i^{\text{grader}} \\
\text{where } 
X_i^{\text{grader}} =
\begin{cases}
+1 & \text{if Autograder} \\
-1 & \text{if Human}
\end{cases}
\end{gathered}
\label{eq:model_autograder}
\end{equation}

where $\phi_i$ is the linear predictor on the latent scale, $\boldsymbol{c}$ is the vector of cutpoints that divide the latent scale into discrete score categories, and the ordered logistic model uses a cumulative logit link function to map from the linear predictor to score probabilities through the relation: $\text{logit}(P(\text{score}_i \leq j)) = c_j - \phi_i.$ The variable $X_i^{\text{grader}}$ encodes the identity of the grader who assigned score $i$ using effect coding.\footnote{Effect coding is a contrast coding scheme used in to represent categorical variables. Each level of the variable is assigned a numerical value such that the resulting coded variables are centered and sum to zero across all groups. This centering allows the intercept to reflect the grand mean (the average of the group means), and the coefficients represent deviations from that mean.} The linear model has two parameters: an intercept $\beta_0$, and a grader effect $\beta_1$. The intercept $\beta_0$ represents the average latent score across both graders, due to the centering imposed by effect coding. The grader coefficient $\beta_1$ captures the effect of the grader on the latent score and quantifies half the difference between the autograder and the human grader. Since the model encodes the autograder as $+1$ and the human as $-1$, the linear predictor shifts depending on the grader identity. These parameters define the location of the latent score, which is then mapped to the observed 1-10 categories via the threshold cutpoints. This shift in the latent score translates to a systematic difference in the probabilities of assigning higher or lower scores. 

To evaluate hypotheses, Florence fits different models to the data and compares their relative fit using statistical criteria such as Leave-One-Out cross-validation (LOO-CV) or the Widely Applicable Information Criterion (WAIC). For instance, here, Florence should compare a model that includes a grader effect (\cref{eq:model_autograder}) against a simpler model without this effect (i.e., $\phi_i = \beta_0$). The former statistical model tests the hypothesis that grader identity affects the assigned score, while the latter one serves as a null model with no grader effect. Model comparison indicates whether adding the grader effect improves model fit, but it does not guarantee that the grader parameter itself is significant. Therefore, once the best-fitting model (often called ``winning model'') is identified, Florence should inspect the parameter estimates and their credible intervals to confirm if there is clear statistical evidence for the grader effect.

After fitting the statistical models and running a model comparison (\cref{fig:model_selection_Q1} in \cref{sec:model-comparison}), we identify the model with the grader effect (\cref{eq:model_autograder}) as the best-fitting model. We can then examine the estimated effect of interest of this model (right panel of \cref{fig:fig_autograder}).
Since the grader variable is effect coded as $+1$ for the autograder and $-1$ for the human, the plotted difference in latent score between the autograder and the human grader is $2\beta_1$. This difference is defined as \emph{autograder minus human}, so a negative value indicates that the human grader assigns higher scores on the latent scale. We observe a negative difference, with credible intervals that do not include zero, indicating a meaningful difference between the two graders. This confirms that the autograder systematically assigns lower scores than the human grader.

Florence now has clear statistical evidence that the autograder gives scores lower than she does, and she can make informed decisions based on this. She could, for instance, revise the autograder’s prompt to encourage higher scores, or she might proceed with the current setup while keeping in mind that the scores tend to be more conservative than her own. The estimated effects are measured on the model's latent scale, which has arbitrary units that do not directly correspond to the 1-10 scoring scale. To understand the practical magnitude of this difference, Florence could simulate predicted scores from the model to see how the latent scale difference translates to differences in the observed 1-10 scores.\looseness-1

If the observed difference in scores seems small, Florence might also want to assess whether this difference is negligible for practical purposes. One way to do this is by defining a Region of Practical Equivalence (ROPE; \citep{kruschke2018rejecting}), an interval around zero within which effect sizes are considered too small to matter. A commonly used ROPE for standardized effect sizes is from $-0.1$ to $0.1$, which corresponds to approximately $-0.18$ to $0.18$ on the log-odds scale used by logistic models. If the entire credible interval of the grader coefficient falls within this ROPE, Florence can conclude that the autograder and the human grader are practically equivalent in terms of their scoring.

\subsection*{1.2 Integrating the autograder assessment with the research question} 
\phantomsection
\label{sec:q12}

The advantage of the GLM framework is that the evaluation of the autograder can be seamlessly integrated with the research question of interest. Going back to Florence, her goal is to assess an LLM’s ability to answer open-ended questions. Lets imagine that she needs to choose between two LLMs, which we denote by A and B. Using a GLM framework, she can address those two critical questions simultaneously:
\begin{enumerate}
    \item Can my autograder reliably evaluate LLM responses relative to a human annotator?
    \item Is LLM A or LLM B better at answering open-ended questions?
\end{enumerate}

\begin{figure}
  \centering
  \includegraphics[width=0.9\textwidth]{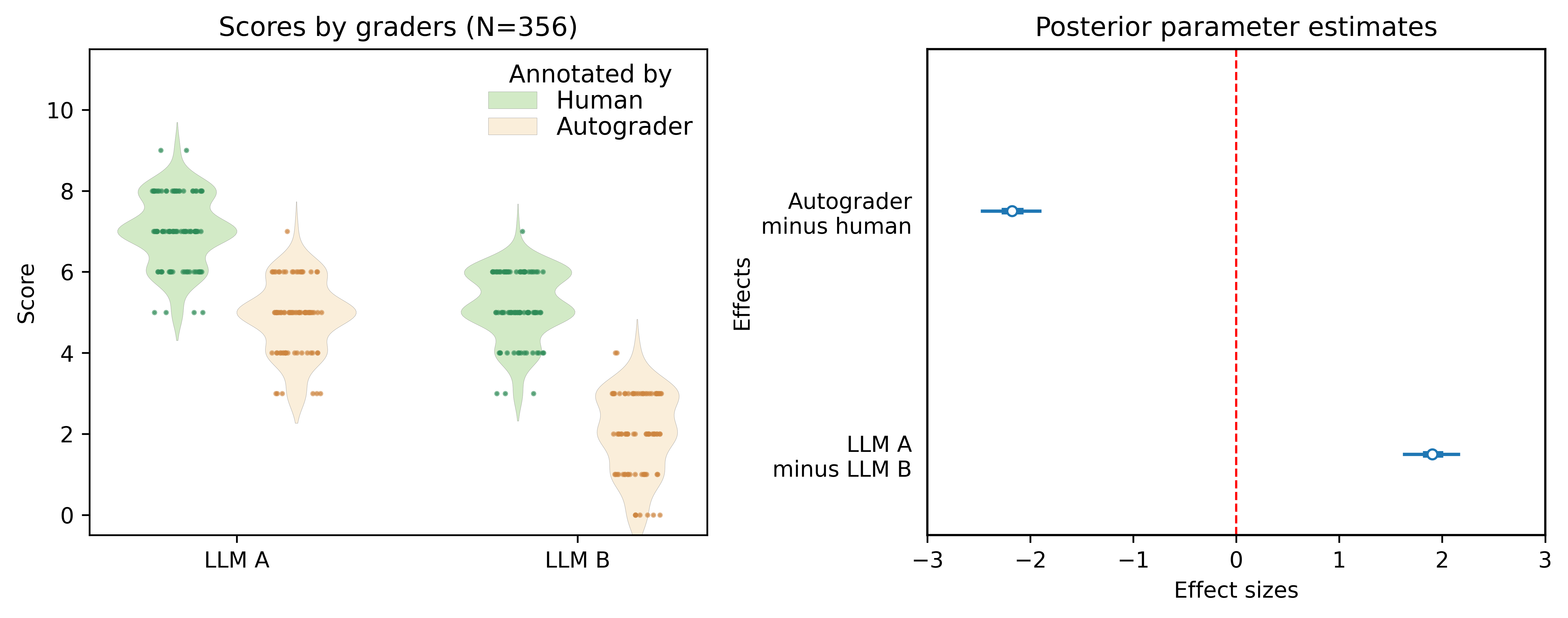}
  \caption{Illustration of how to use Bayesian GLMs to address \hyperref[sec:q12]{Question 1.2} (quantifying the mean difference between scores assigned by an autograder and a human grader while evaluating LLMs) using simulated data. Left panel: Violin plot of simulated scores for LLM-generated answers graded by a human expert (Florence in the example) and an autograder. Right panel: Posterior distributions of estimated effects. The horizontal blue lines represent 95\% credible intervals. The dashed red vertical line indicates a null effect ($\beta = 0$). The coefficient for autograder minus human is negative, with a credible interval that does not include zero, indicating that the autograder tends to assign lower scores. The coefficient for LLM A minus LLM B is positive, suggesting that LLM A receives higher scores than LLM B on average.}
  \label{fig:fig_autograder_llm}
\end{figure}

Now suppose that the $N = 100$ graded responses (each graded by both Florence and her autograder) are in fact 50 unique items, with each item answered once by LLM A and once by LLM B. The data might look like the left panel of \cref{fig:fig_autograder_llm}. To answer the two questions above, we can fit a regression model with an intercept $\beta_0$, which represents the overall average latent score, and a coefficient $\beta_1$ that quantifies the effect of the grader, as before (\cref{eq:model_autograder}). In addition, we include a coefficient $\beta_2$ to quantify the effect of LLM A versus LLM B on the score.

\begin{equation}
\begin{gathered}
\text{score}_i \sim \text{OrderedLogistic}(\phi_i, \boldsymbol{c}) \\
\phi_i = \beta_0 + \beta_1 \cdot X_i^{\text{grader}} + \beta_2 \cdot X_i^{\text{LLM}} \\
\text{where } 
X_i^{\text{grader}} =
\begin{cases}
+1 & \text{if Autograder} \\
-1 & \text{if Human}
\end{cases},
\quad
X_i^{\text{LLM}} =
\begin{cases}
+1 & \text{if LLM A} \\
-1 & \text{if LLM B}
\end{cases}
\end{gathered}
\label{eq:model_autograder_llm}
\end{equation}

 The variables $X_i^{\text{grader}}$ and $X_i^{\text{LLM}}$ encode, respectively, the identity of the grader and the LLM associated with score $i$, using effect coding. Each variable takes a value of $+1$ or $-1$ to distinguish between the two levels (e.g., autograder vs.\ human, LLM A vs.\ LLM B). This coding imposes a sum-to-zero constraint on each effect, so that $\beta_1$ and $\beta_2$ represent half the difference in latent score between levels of the corresponding variable.

The coefficients $\beta_1$ and $\beta_2$ allow us to answer the two questions above in a single statistical model. After fitting the model, we can make inferences based on the effect sizes represented by the coefficients (right panel of \cref{fig:fig_autograder_llm}). We observe two things:
\begin{enumerate}
    \item Similarly to before, the \emph{autograder minus human} effect ($\beta_1 - (-\beta_1) = 2\beta_1$) is negative, which indicates that the autograder gives lower scores than the human expert.
    \item In the same way, the \emph{LLM A minus LLM B} effect ($\beta_2 - (-\beta_2) = 2\beta_2$) is positive, which indicates that LLM A receives higher scores on average than LLM B.
\end{enumerate}

In both cases, the credible intervals exclude zero, providing strong evidence for these effects.

From this analysis, Florence can conclude that while there exists a discrepancy between human and autograder scoring, LLM A is significantly better than LLM B at answering open-ended questions. She can therefore confidently select LLM A for her task, and in parallel, start improving the autograder with the insight that it tends to assign lower scores than human graders. This illustrates how a GLM framework can be used to simultaneously address a specific research question and assess the quality of autograders.

\section*{Question 2: Do autograders favour their own generation?}
\phantomsection
\label{sec:q2}
Recent literature has raised concerns that autograders may demonstrate self-bias, a tendency to assign better scores to outputs generated by the same base model \citep{panickssery-2024-recognize,liu-etal-2024-llms-narcissistic,koo-etal-2024-biases} or outputs from models vs.\ humans \citep{liu-etal-2023-geval}, which would naturally compromise evaluation automation efforts. 
Similarly to above, with a GLM framework, we can seamlessly integrate this analysis into a specific research question and quantify self-bias for a particular application.

\begin{figure}
  \centering
  \includegraphics[width=1.0\textwidth]{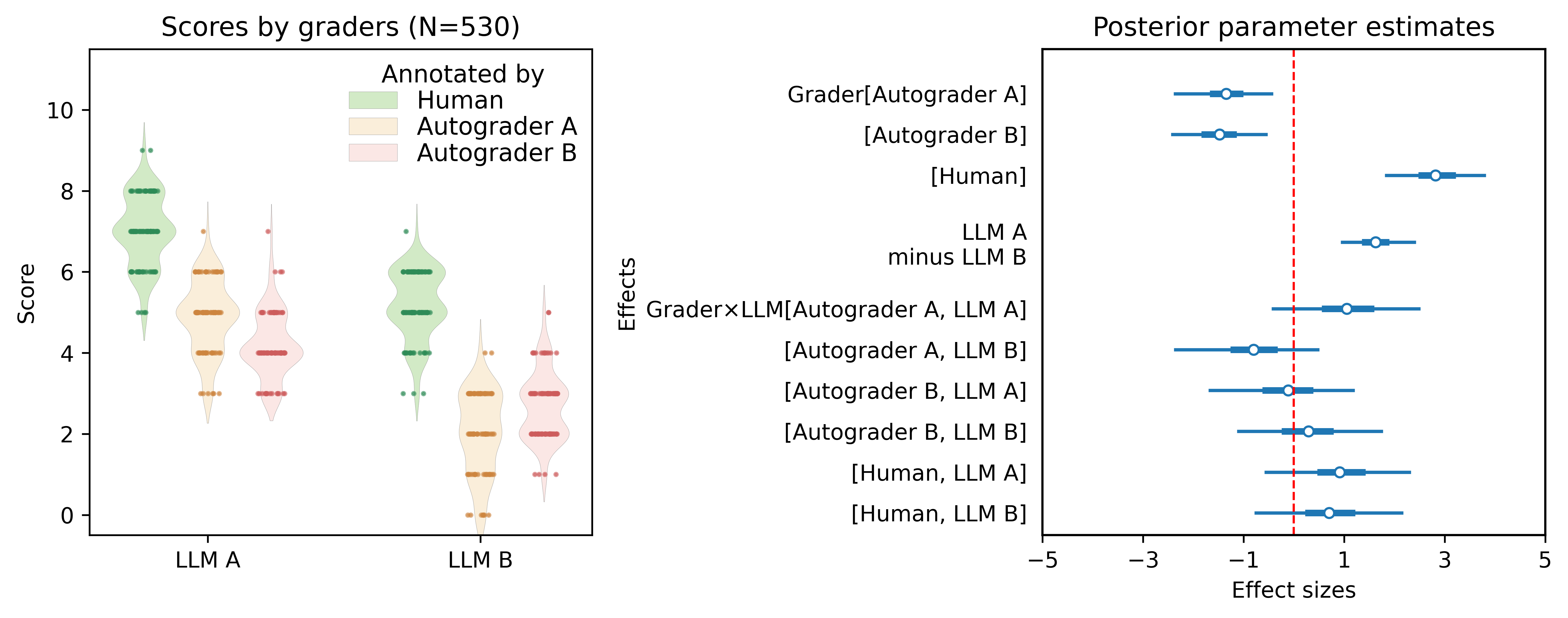}
  \caption{Illustration of how to use Bayesian GLMs to address \hyperref[sec:q2]{Question 2} (Do autograders favour their own generation?) using simulated data. Left panel: Violin plot of simulated scores for LLM-generated answers by two LLMs (LLM A and LLM B). The scores were given by a human expert (green) and two autograders (yellow and red). Right panel: Posterior distributions of estimated effects from the GLM. The horizontal blue lines represent 95\% credible intervals, and the dashed red vertical line indicates a null effect ($\beta = 0$). The grader effect ($\beta_1$) shows how each grader deviates from the average score across all graders and LLMs. The LLM effect ($\beta_2$) is positive, indicating that LLM A generally receives higher scores than LLM B. The grader–LLM terms ($\beta_3$) represent a set of parameters (one for each grader–LLM combination) estimated using index-based coding. These parameters are not traditional interaction effects (e.g., a single $\beta_3 \cdot X_{\text{grader}} \cdot X_{\text{LLM}}$ term), but instead allow for direct comparison between specific combinations. The difference in estimated effects for Autograder A on LLM A versus Autograder B on LLM A reveals a non-overlapping contrast, consistent with grader-specific scoring preferences. Together, these results suggest that each autograder favours the LLM it was developed on, indicative of a systematic self-bias.}
  \label{fig:fig_autograder_llm_graders}
\end{figure}

Going back to Florence, she recently learned about potential self-bias of autograders and is concerned that her autograder (from model family A) might unfairly favour outputs from LLM A (also from model family A). To assess whether such self-bias exists, she uses a second autograder (from model family B). She wants to investigate whether responses from LLM A receive higher scores when graded by the autograder A compared to when graded by autograder B (and vice versa). The resulting data are shown in the left panel of \cref{fig:fig_autograder_llm_graders}.  

To answer the question about self-bias, we extend the previous model (\cref{eq:model_autograder_llm}) by adding a term that captures whether specific graders systematically favour outputs from specific LLMs. This is implemented as a set of grader–LLM interaction effects\footnote{Strictly speaking, this is not a single interaction effect (e.g., $\beta_3 \cdot X^{\text{grader}} \cdot X^{\text{LLM}}$), but a set of parameters estimated independently using index-based coding (where each grader–LLM combination has a unique integer index). This allows direct comparisons across specific combinations rather than relying on a single coefficient.}, denoted by $\beta_3$.

\begin{equation}
\begin{gathered}
\text{score}_i \sim \text{OrderedLogistic}(\phi_i, \boldsymbol{c}) \\
\phi_i = \beta_0 + \beta_1(X_i^{\text{grader}}) + \beta_2 \cdot X_i^{\text{LLM}} + \beta_3(X_i^{\text{grader}}, X_i^{\text{LLM}})
\end{gathered}
\label{eq:model_autograder_llm_interaction}
\end{equation}

where $\phi_i$ is the linear predictor on the latent scale, $\boldsymbol{c}$ is a vector of cutpoints, and $X_i^{\text{grader}}$ and $X_i^{\text{LLM}}$ identify the grader and LLM assigned to observation $i$. As we now have more than two graders, the main effect $\beta_1$ is a vector of coefficients representing each grader’s deviation from the grand mean score, estimated using effect coding. The LLM variable still has only two levels and is binary-coded as before (see \cref{eq:model_autograder_llm}), so $\beta_2$ is a scalar coefficient applied to $X_i^{\text{LLM}} \in \{-1, +1\}$. The interaction term $\beta_3$ is a matrix of parameters estimated using index-based coding, with one distinct coefficient independently estimated for each grader–LLM combination. Unlike effect coding, which estimates deviations from a grand mean (i.e., the average across all grader–LLM combinations), and dummy coding, which estimates differences from a reference category, index-based coding directly estimates each combination independently relative to the intercept. This allows for direct interpretation of each pair (e.g., how Autograder scores LLM A compared to the intercept) and simplifies comparisons across combinations.

After selecting the best-fitting model using model comparison techniques (\cref{fig:model_selection_Q2} in \cref{sec:model-comparison}), we examine the estimated effects (right panel of \cref{fig:fig_autograder_llm_graders}). The interaction parameters $\beta_3$ include a positive effect for the combination of Autograder A and LLM A, indicating that Autograder A tends to assign higher scores to LLM A relative to the overall baseline (i.e., the intercept). Conversely, the interaction term for Autograder B and LLM A is negative, suggesting that Autograder B gives lower scores to LLM A compared to the same baseline. While these individual patterns show grader-specific scoring tendencies, the key evidence for self-bias emerges from directly contrasting how Autograder A scores LLM A against how Autograder A scores LLM B. Looking at these specific effects, we see a difference in their posterior means suggesting a tendency toward self-bias (although, because their credible intervals partially overlap, additional data would be necessary for a more definitive conclusion).

With these findings, Florence confidently answers her main question (LLM A performs better on open-ended questions), uncovers that the autograders assign lower scores than she does, and identifies a potential self-bias in their evaluations. To address this bias, she could choose to use independently developed autograders in the future, or simply factor these biases into her interpretation of results.

\section*{Question 3: Do autograders differ systematically from human experts?}
\phantomsection
\label{sec:q3}

\begin{figure}
  \centering
  \includegraphics[width=0.9\textwidth]{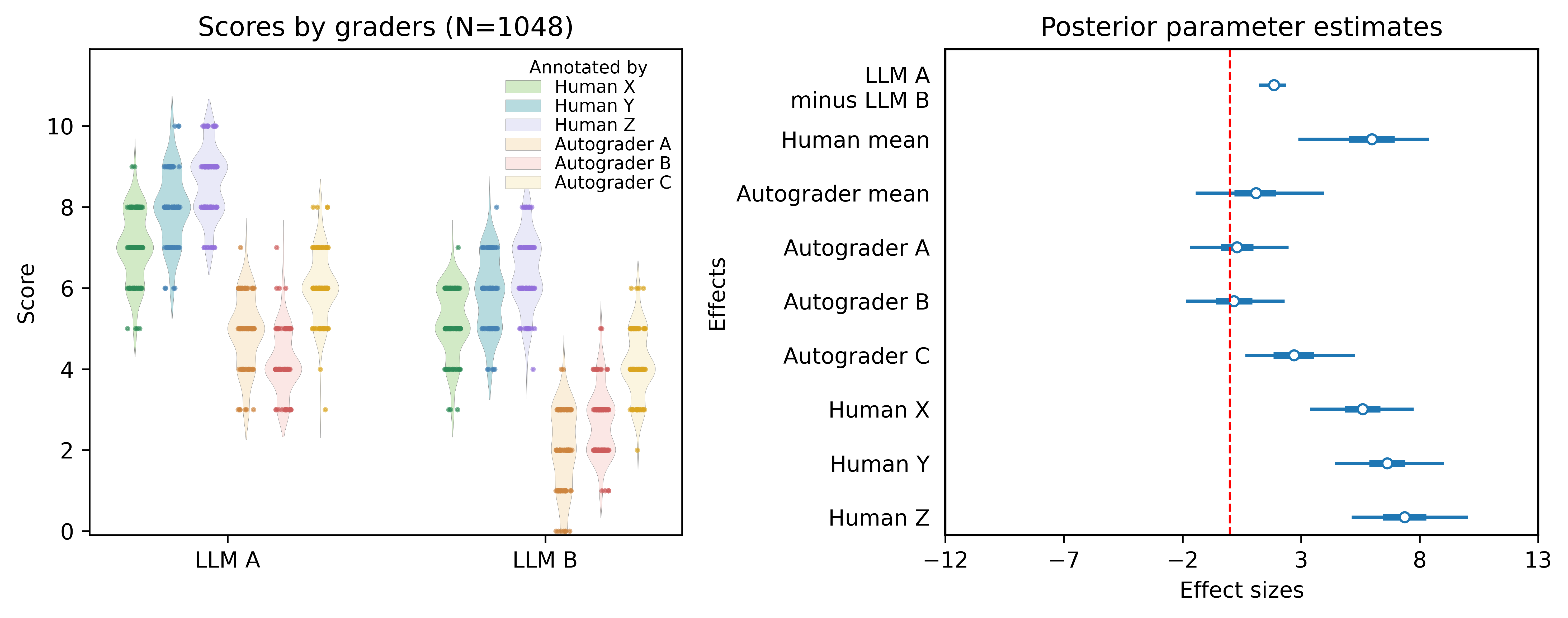}
  \caption{Illustration of how to use Bayesian GLMs to address \hyperref[sec:q3]{Question 3} (Do autograders differ systematically from human experts?) using simulated data. Left panel: Violin plot of simulated scores for LLM-generated answers from two models (LLM A and LLM B), as graded by multiple human experts (green) and autograders (yellow and red). Right panel: Posterior distributions of estimated effects from the hierarchical model. The horizontal blue lines represent 95\% credible intervals, and the dashed red vertical line indicates a null effect ($\beta = 0$). Individual grader effects show how each grader deviates from their respective group-level average (human or autograder). The plotted effect for LLM A minus LLM B represents $2\beta_2$, capturing the full latent score difference between the two models under effect coding. Group-level means for human and autograder graders ($\mu_{\text{graderType}}$) indicate that, on average, human graders assign higher scores than autograders.}
  \label{fig:fig_autograder_llm_graders_type}
\end{figure}

In practice, Florence will probably not have the time (nor the motivation) to grade even N=100 responses herself. Instead, she will ask a few of her colleagues to help grade some of the responses. Instead of only having grades from one human grader, she will now have grades from multiple humans, and can and can assess their individual scoring patterns (Human X, Y and Z in the left panel of \cref{fig:fig_autograder_llm_graders_type}.) Additionally, because she is aware of potential autograder biases, she might want to try other autograders, for instance LLMs from different model families, or LLMs from the same family but using different prompts.  Just like with the humans, she will now have grades from different autograders that she can examine individually (Autograder A, B and C  in the left panel of \cref{fig:fig_autograder_llm_graders_type}.). 

In terms of analysis, she could account for each grader individually, similarly to the $\beta_1$ effect in \cref{eq:model_autograder_llm} (visualised in the right panel of \cref{fig:fig_autograder_llm_graders}). The problem with that method is that the number of parameters increases with the number of graders, potentially complicating interpretation.  

Instead, she might prefer to treat human graders and autograders as two broader groups, while considering the human scores as providing a form of ground truth, and assess whether autograder scores, on average, differ systematically from those human scores. If a meaningful difference between the two groups is observed, it could suggest the need to revise how the autograders are instructed or used. 

To capture such a group-level difference, we can define what is known as a hierarchical GLM (cf. \cref{eq:model_hierarchical}). In this model, each individual grader has their own average scoring tendency ($\beta_{\text{grader}_i}$), but these grader-specific effects are drawn from a group-level distribution depending on their type (human or autograder). This approach allows us to estimate separate group-level means for human graders and autograders, which can then be directly compared to assess systematic differences between the two groups. It also captures individual-level deviations, which reflect how each grader differs from the average behaviour of their group. This makes it possible to detect outliers or particularly lenient or strict graders. By sharing information across graders of the same type, a property known as partial pooling, the statistical model makes more efficient use of limited data, which is particularly helpful when some graders have relatively few observations.

\begin{equation}
\begin{gathered}
\text{score}_i \sim \text{OrderedLogistic}(\phi_i, \boldsymbol{c}) \\
\phi_i = \beta_0  + \beta_1(X_i^{\text{grader}}) + \beta_2 \cdot X_i^{\text{LLM}} \\
\beta_1(X_i^{\text{grader}}) \sim \mathcal{N}(\mu_{\text{graderType}_i}, \sigma^2_{\text{graderType}_i}) \\
\mu_{\text{graderType}_i} \sim \mathcal{N}(0, 3) \\
\sigma^2_{\text{graderType}_i} \sim \text{HalfCauchy}(1) \\
\text{with } \boldsymbol{c} \text{ as vector of cutpoints}
\end{gathered}
\label{eq:model_hierarchical}
\end{equation}

where $\phi_i$ is the linear predictor on the latent scale, and $\boldsymbol{c}$ is a vector of cutpoints that divide the latent scale into discrete score categories. The intercept $\beta_0$ represents the grand mean, and $\beta_2$ is a scalar coefficient applied to $X_i^{\text{LLM}} \in \{-1, +1\}$, which indicates whether the response was generated by LLM A or LLM B. Here, $\beta_1(X_i^{\text{grader}})$, unlike before, represents the effect of the individual grader who assigned score $i$, and is drawn from a group-level distribution based on grader type (human or autograder). Specifically, $\beta_1(X_i^{\text{grader}}) \sim \mathcal{N}(\mu_{\text{graderType}_i}, \sigma^2_{\text{graderType}_i})$, where $\mu_{\text{graderType}_i}$ represents the average score tendency for each grader type, and $\sigma^2_{\text{graderType}_i}$ captures variability within each type. The prior distributions for the group-level means and variances ($\mu_{\text{graderType}}$ and $\sigma^2_{\text{graderType}}$) are specified in \cref{sec:priors}. This hierarchical structure enables the model to estimate both the average difference between human and autograder scores and the variation among individual graders.

To formally assess whether this group-level difference exists, Florence should of course compare this hierarchical approach against the simpler flat model (cf. \cref{fig:model_selection_Q3} in \cref{sec:model-comparison} for a model comparison). Here we select the hierarchical model to demonstrate how to examine both group-level differences between grader types and individual grader characteristics.

In the right panel of \cref{fig:fig_autograder_llm_graders_type} we see the individual grader effects ($\beta_1$; Autograder A–C and Human X–Z in the plot) that provide information about how each grader scores on average. The group-level means for human and autograders  ($\mu_{\text{graderType}}$; human mean and Autograder mean in the plot) allow us to estimate the average difference in scoring tendencies between the two groups. The results show that human graders systematically assign higher scores than autograders, regardless of individual differences.

Using this method, Florence can confidently conclude that there is a general tendency for humans to give higher scores than autograders, and she can interpret her main results with this in mind. Additionally, she can visualise individual-level differences and make informed decisions. For example, she might observe that Autograder C produces scores that are more closely aligned with those of the human graders. If consistency with human judgment is a key objective, she may choose to use this autograder in future evaluations.

\section*{Question 4: How do scores differ at an item level?}
\phantomsection
\label{sec:q4}

\begin{figure}
  \centering
  \includegraphics[width=0.9\textwidth]{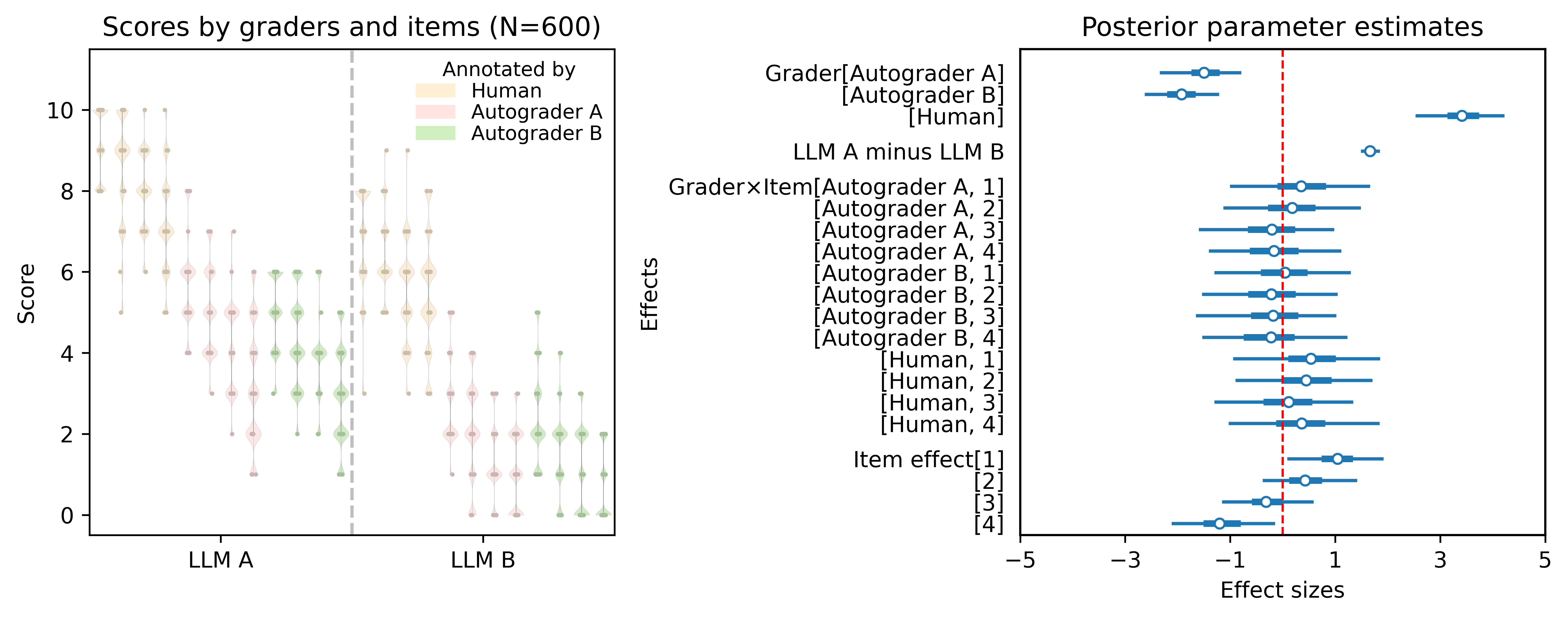}
  \caption{Illustration of how to use Bayesian GLMs to address \hyperref[sec:q4]{Question 4} (How do scores differ at an item level?) using simulated data. Left panel: Violin plot of simulated scores for each item (1–4), grouped by LLM and grader identity. Each cell shows the distribution of scores assigned by a given grader to responses from a particular model on a given item. Right panel: Posterior distributions of estimated effects from the item-level GLM (\cref{eq:model_item_grader}). The plot shows main effects for grader and LLM identity (top), item main effects (bottom), and grader–item interactions (middle). Horizontal blue lines represent 95\% credible intervals, and the dashed red vertical line indicates a null effect ($\beta = 0$). Item 1 has a strong positive effect, suggesting it consistently receives higher scores. In contrast, Item 4 has a negative effect, indicating that it receives lower scores. Grader–item interaction terms are small and uncertain, indicating no evidence of systematic grader disagreement on specific items.}
  \label{fig:fig_autograder_item}
\end{figure}

Florence now becomes interested in whether variation arises at the level of individual evaluation items (i.e., open-ended questions). She wonders whether some items consistently receive higher or lower scores, and whether graders agree more on certain items than others.

To answer these questions, she needs repeated responses for the same items. Until now, we have assumed that each data point corresponds to a different item, so lets instead imagine that Florence’s dataset consists of four items (for simplicity), with each model answering each item 25 times. This results in 50 responses per item (25 per model); each grader scores all of the attempts.  The data split by items can be seen in the left panel of \cref{fig:fig_autograder_item} (different items are represented by violin plots of the same colour).

To answer Question 4, we extend \cref{eq:model_autograder_llm} by including two additional terms. 
The first term, $\beta_3(X_i^{\text{item}})$, accounts for a main effect of items, capturing whether some items receive systematically higher or lower scores. 
The second term, $\beta_4(X_i^{\text{grader}}, X_i^{\text{item}})$, represents a grader–item interaction, allowing us to test whether particular graders behave differently on specific items.

\begin{equation}
\begin{gathered}
\text{score}_i \sim \text{OrderedLogistic}(\phi_i, \boldsymbol{c}) \\
\phi_i = \beta_0 + \beta_1(X_i^{\text{grader}}) + \beta_2 \cdot X_i^{\text{LLM}} + \beta_3(X_i^{\text{item}}) + \beta_4(X_i^{\text{grader}}, X_i^{\text{item}})
\end{gathered}
\label{eq:model_item_grader}
\end{equation}

Here, $\phi_i$ is the linear predictor on the latent scale, and $\boldsymbol{c}$ is the vector of cutpoints that divide the latent scale into ordered categories. The term $\beta_1(X_i^{\text{grader}})$ captures the main effect of each grader, and $\beta_2$ models the effect of LLM identity (e.g., whether the response was produced by LLM A or B). 
As mentioned above, the new term $\beta_3(X_i^{\text{item}})$ represents the main effect of each item, which captures whether some questions tend to receive higher or lower scores overall. The final term, $\beta_4(X_i^{\text{grader}}, X_i^{\text{item}})$, captures grader–item interactions and is implemented in the same way as the interaction term in \cref{eq:model_autograder_llm_interaction}: rather than a single multiplicative term, this is a set of coefficients (one for each grader–item combination) estimated using dummy coding without a sum-to-zero constraint. This allows us to directly compare individual combinations and detect whether certain graders are more lenient or harsh on specific items. As before, all main categorical effects (grader, item) are encoded using effect coding, so that the resulting coefficients reflect deviations from the overall mean. 

After fitting the model, Florence inspects the estimated effects (right panel of \cref{fig:fig_autograder_item}). She focuses on two aspects: 
\begin{enumerate}
    \item The main effect of item ($\beta_3$ in \cref{eq:model_item_grader}), observing that some items consistently receive higher or lower scores. For example, Item 1 shows a notably positive main effect, indicating that graders tend to assign higher scores to responses for this item. Conversely, Item 4 exhibits a negative effect, implying that graders give lower scores on this item. Overall, this suggests that Item 1 is easier to answer, while Item 4 is more challenging. 
    \item The grader–item interaction term ($\beta_4$ in \cref{eq:model_item_grader}), which helps identify whether graders differ more on specific questions. Most interaction estimates are small and their credible intervals overlap with zero, indicating weak evidence that specific graders behave differently on individual items. 
\end{enumerate}

From these results, Florence concludes that while some items appear easier than others, grader disagreement is not concentrated on any particular question. 
However, she still wants to understand how consistently graders agree overall across all items.
A common way to quantify this is to compute an inter-rater agreement statistic such as Krippendorff’s $\alpha$ \citep{tam2024framework,bavaresco-2025-instead}, which measures the consistency of annotations across (more than 2) raters.

\begin{figure}
  \centering
  \includegraphics[width=0.9\textwidth]{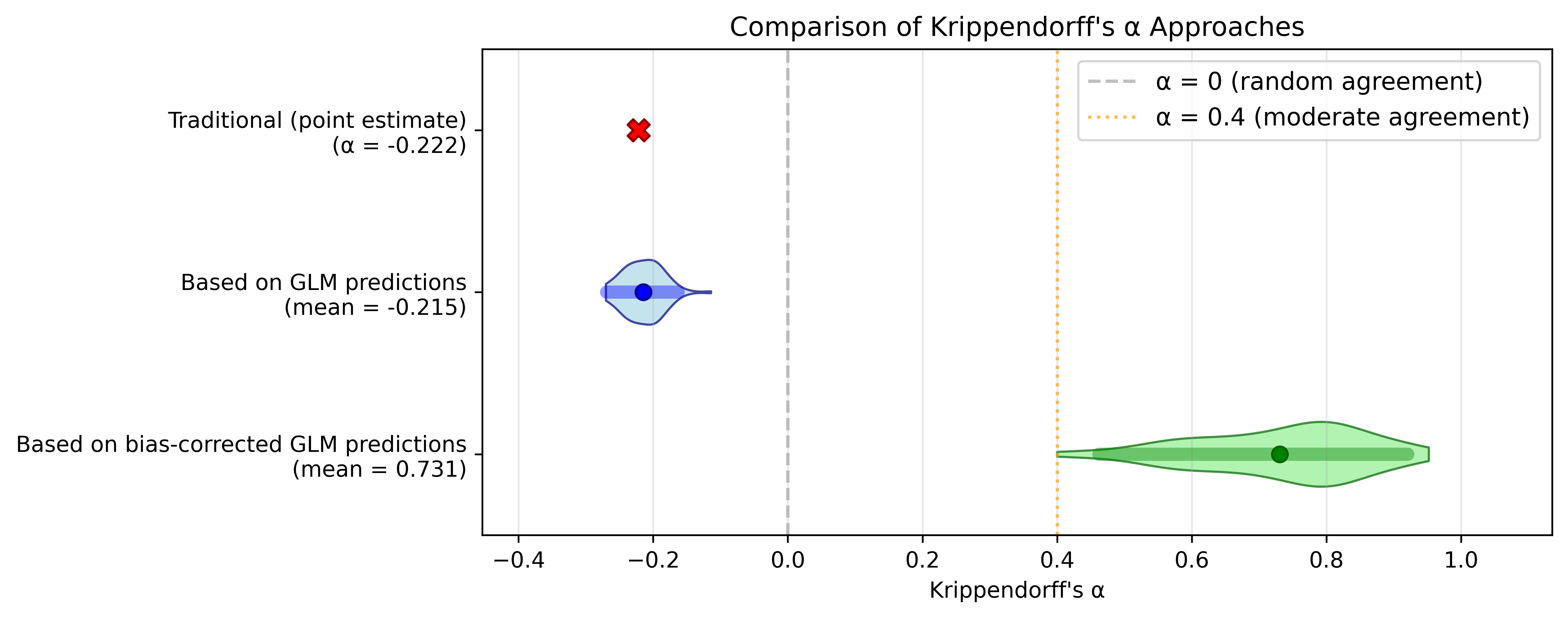}
  \caption{Posterior distributions of Krippendorff’s $\alpha$ under different modeling assumptions. The red cross shows the traditional $\alpha$ computed directly from the observed scores, suggesting strong disagreement between graders. The blue distribution shows $\alpha$ values estimated from posterior simulations of the fitted GLM, incorporating uncertainty in the predictions. The green distribution shows $\alpha$ values after removing the main effect of grader identity, revealing what agreement might look like in a counterfactual scenario where graders do not differ systematically in scoring scale. The wider spread reflects greater uncertainty under this assumption, as individual grader variation is no longer explained.}
\label{fig:fig_alpha}
\end{figure}

Using the posterior probabilities from the fitted GLM, Florence can simulate scores and compute a distribution over $\alpha$ values, giving her not only a point estimate but also a full uncertainty interval around inter-rater agreement. This is shown in \cref{fig:fig_alpha}, where the red cross denotes the standard Krippendorff $\alpha$ computed directly from the observed scores, and the blue distribution represents the GLM-based estimate. This is a significant improvement over traditional methods, which typically return a single value and do not account for sampling variability or data sparsity.

However, while helpful, such agreement metrics have important limitations: they do not distinguish between disagreement due to noise and disagreement due to systematic bias. In \cref{fig:fig_alpha}, both the traditional and GLM-based $\alpha$ estimates are below zero, suggesting substantial disagreement. But this does not tell us why graders disagree. To answer this question, Florence turns back to the right panel of \cref{fig:fig_autograder_item}, where she observes that human graders have a strong positive main effect relative to autograders, meaning they consistently assign higher scores. She also notes that while items differ in average difficulty, the grader–item interaction effects are close to zero, indicating that graders do not differ much in how they score specific items.

Together, these two observations suggest that the low $\alpha$ is not due to random noise, but to a systematic scoring bias across grader types. To illustrate this, Florence goes one step further: for each predicted score, she subtracts the estimated main effect of the corresponding grader (i.e., their bias term from the model) from the linear predictor before mapping it to a categorical score. This yields a set of bias-adjusted predictions that reflect how graders might score if they did not differ systematically in their baseline scoring. In essence, Florence is constructing a counterfactual scenario: ``What would agreement look like in the absence of systematic differences among graders?''

She then samples from these bias-adjusted posterior predictions and recomputes Krippendorff’s $\alpha$. The resulting agreement score posterior, shown as the green distribution in \cref{fig:fig_alpha}, is substantially higher, indicating that most of the disagreement was driven by consistent shifts in scoring behaviour rather than inconsistency in judgment. Notably, the green distribution is also wider than the others. This is most likely because the model has removed a major source of predictable variation, i.e., the average bias associated with each grader, but has not explicitly accounted for individual differences in scoring behaviour. As a result, variability that was once explained by grader identity now becomes unexplained, increasing uncertainty in the prediction. 

In conclusion, this framework allows Florence to go beyond traditional agreement metrics and understand the source of disagreement. She can still compute familiar statistics such as Krippendorff’s $\alpha$, Cohen’s $\kappa$, or Kendall’s $\tau$, but now using posterior predictions from the GLM. These predictions account for covariates in the data and yield credible intervals for each metric, enabling more robust and interpretable comparisons across grader types, task formats, and evaluation settings.

\section*{Question 5: Do autograders favour longer outputs?}
\phantomsection
\label{sec:q5}

So far, we have focused on evaluation settings where graders assign absolute scores. However, many LLM evaluations rely on pairwise comparisons, where graders are asked to choose which of two outputs better satisfies a target criterion (e.g., correctness). The same statistical modeling framework can be applied in this setting, with the outcome modeled as a binary preference. We use such a setup to illustrate how pairwise comparisons can be modeled and to examine length bias, which has often been observed in pairwise evaluation settings. Of course, length bias can also be captured in absolute score setups similarly to other biases in previous sections.

\begin{figure}
  \centering
  \includegraphics[width=0.9\textwidth]{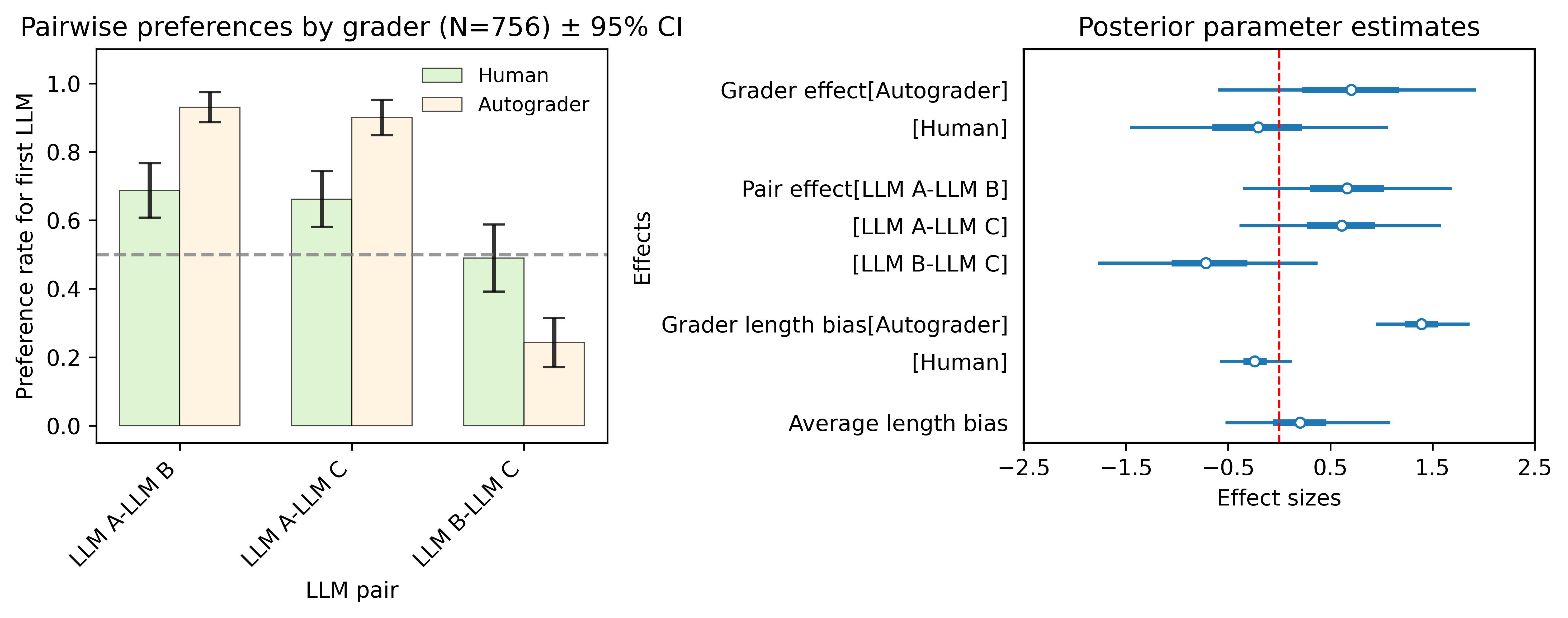}
  \caption{Illustration of how to use Bayesian GLMs to address \hyperref[sec:q5]{Question 5} (Do autograders favour longer outputs?) using simulated data. Left panel: Proportion of pairwise preferences across three LLM pairs (A vs.\ B, A vs.\ C, B vs.\ C). Preference rate values represent the fraction of cases where the first listed model was selected over the second. Error bars represent 95\% confidence intervals. Right panel: Posterior distributions of estimated effects from the GLM for pairwise comparisons (log-odds scale). Horizontal blue lines represent the 95\% credible intervals, and the dashed red vertical line indicates a null effect ($\beta$=0). The pair effect terms represent the relative preference between specific pairs of LLMs, indicating which LLM is generally preferred. The grader length bias terms quantify each grader's sensitivity to token-length differences when making choices. A positive length bias indicates a preference towards longer outputs.
  }
  \label{fig:fig_autograder_pairwise}
\end{figure}

Let's imagine that Florence wants to compare the quality of outputs generated by three different LLMs. She chooses a pairwise evaluation format, where each grader (e.g., herself or an autograder) is repeatedly shown two responses to the same prompt - each generated by a different LLM - and must select the better response. An example of such data can be seen in the left panel of \cref{fig:fig_autograder_pairwise}. Each bar represents a pairwise comparison (e.g., ``LLM A vs.\ LLM B''), and its height reflects how frequently the first listed model (e.g., LLM A) was chosen. To model this, we can use a binomial GLM with a logit link function. The outcome variable $y_i$ indicates whether the first model in the pair was chosen ($y_i = 1$) or not ($y_i = 0$), and we include a categorical effect in the model to denote the LLM pair being compared.

Florence's younger brother, always up-to-date with ML controversies, recently told her that some autograders may systematically prefer longer outputs even if those outputs are not of higher quality, a phenomenon commonly referred to as length bias \citep{zheng2023judging, dubois2024lengthcontrolled}. To capture this bias, she adds a continuous predictor capturing the token-length difference between the two outputs. As there are two graders (herself and the autograder), which might have different biases, she computes one such predictor per grader. To test the existence of the length bias formally, she compares the model with and without this term (cf. \cref{fig:model_selection_Q5} in \cref{sec:model-comparison}). For demonstration purposes, let's look at the model with grader-specific length bias here:

\begin{equation}
\begin{gathered}
y_i \sim \text{Binomial}(1, p_i) \\
\text{logit}(p_i) = \beta_0 + \beta_1(X_i^{\text{LLM pair}}) + \beta_2(X_i^{\text{grader}}) + \beta_3(X_i^{\text{grader}})\cdot X_i^{\text{lengthDiff}} \\
\beta_3(X_i^{\text{grader}}) \sim \mathcal{N}(\mu_{\text{lengthDiff}}, \sigma_{\text{lengthDiff}}^2) \\
\mu_{\text{lengthDiff}} \sim \mathcal{N}(0, 0.5) \\
\sigma_{\text{lengthDiff}} \sim \text{HalfNormal}(1.0) \\
\end{gathered}
\label{eq:model_pairwise_hierarchical}
\end{equation}

where $y_i$ is a binary outcome indicating whether the first-listed LLM was preferred. The intercept $\beta_0$ is the overall tendency to prefer the first-listed model, $\beta_1(X_i^{\text{LLM pair}})$ captures pair-specific preferences (e.g., ``LLM A vs. LLM B'') and $\beta_2(X_i^{\text{grader}})$ captures each grader's overall tendency to prefer the first-listed model. The predictor $X_i^{\text{lengthDiff}}$ is the token-length difference between the two responses. The grader-specific slope coefficient $\beta_3(X_i^{\text{grader}})$ quantifies how sensitive each grader is to length differences and is drawn from a hierarchical distribution with mean $\mu_{\text{lengthDiff}}$ and standard deviation $\sigma_{\text{lengthDiff}}$. Positive values of $\mu_{\text{lengthDiff}}$ indicate a preference for longer outputs. The hierarchical structure captures both the average length bias across graders and the variability among individual graders.

Importantly, once we have estimated the probability of choosing an LLM over another, we can compare these probabilities across pairs. This allows to identify rational (transitive) and irrational (intransitive) patterns of decision-making, such as cyclic dependencies (e.g., preferring A over B, B over C, but C over A). Such intransitivities exist in LLM evaluations \citep{xu2025investigating}. Traditional models like the Bradley–Terry model implicitly assume transitivity and thus cannot capture these cycles. Recent approaches have proposed either removing intransitivities from datasets \citep{yu2025elspr} or explicitly quantifying them \citep{zhang2025beyond, liu2024aligning, zhao2024measuring}. GLMs naturally capture these intransitivities alongside grader biases and differences in LLM performance.

Inspecting the estimated effects in the right panel of \cref{fig:fig_autograder_pairwise}, we observe a positive effect for the grader-specific length bias parameter ($\beta_3$), particularly for the autograder. From this, Florence can infer that the autograder is more likely to select longer outputs, irrespective of their intrinsic quality. In other words, there appears to be an implicit association between output length and perceived correctness in the autograder's judgments. By explicitly quantifying such biases within the model, Florence can more reliably interpret differences in LLM rankings. For example, if LLM A wins most comparisons but consistently produces longer outputs, Florence might question: ``Is LLM A genuinely better, or simply more verbose? Can I really trust the autograder’s judgements?''. Additionally, by examining the estimated LLM pair parameters ($beta_1$), she can verify whether the observed preferences follow a consistent ranking or if there are intransitive (cyclical) patterns. Here, the estimated parameters indicate a consistent ordering: LLM A tends to be preferred to LLM B and LLM C, and LLM B tends to be preferred to LLM C. This integrated statistical framework empowers her to disentangle and quantify these systematic biases and assess preference consistency, leading to deeper and more reliable conclusions.

\section*{Conclusion}

In this paper, we introduced a statistical framework for evaluating autograders using Bayesian GLMs. By jointly modelling the evaluation outcome and the scoring process, this approach enables researchers to assess both LLM performance and autograder behaviour within a single analysis. Through a series of examples, we followed
the journey of a fictional researcher toward evaluating autograders. We used simulated data throughout to explore a wide range of evaluation questions and settings, and to illustrate key modelling principles in a controlled and reproducible way. 

Specifically, we showed how this framework can be used to compute richer, uncertainty-informed estimates of inter-rater agreement that, unlike traditional metrics, help us understand the sources of rater disagreement. We also demonstrated how it can quantify various types of biases (e.g., self-bias and length bias), capture individual-level differences both among graders and  items, and improve the estimation of group-level trends through hierarchical modelling. Additionally, to highlight the method's flexibility, we showed how it can easily be adapted to different evaluation formats, including absolute scoring and pairwise comparisons. 

The examples presented are by no means exhaustive. Many other applications and extensions of GLMs are possible, and we hope this work provides a clear and practical starting point for researchers seeking to adapt the framework to their own evaluation scenarios. To support practical adoption, we have summarised common evaluation questions and their implementation in a GLM setup in \cref{tab:glm_questions}. All models presented in this paper are implemented in the open-source \href{https://github.com/UKGovernmentBEIS/hibayes}{HiBayes package}, and reproducible notebooks for all examples are available in a public \href{https://github.com/magda-dubois/skewed-score}{repository}. 

\newpage
\bibliographystyle{plainnat}
\bibliography{references}

\newpage
\appendix

\section{Priors}
\label{sec:priors}
Below are the priors used across the models described in this paper. They were selected to reflect weakly informative assumptions about effect sizes and score thresholds.
\begin{itemize}
  \item Intercept: $\beta_0 \sim \mathcal{N}(0, 1)$
  \item Main effects of grader: $\beta^{\text{grader}} \sim \mathcal{N}(0, 1)$
  \item Main effects of LLM: $\beta^{\text{LLM}} \sim \mathcal{N}(0, 1)$
  \item Interaction effects: $\beta^{\text{interaction}} \sim \mathcal{N}(0, 1)$
   \item Group-level mean for grader type: $\mu_{\text{graderType}} \sim \mathcal{N}(0, 3)$
  \item Group-level standard deviation: $\sigma^2_{\text{graderType}} \sim \text{HalfCauchy}(1)$
  \item First cutpoint: $c_1 \sim \mathcal{N}(-4.0, 0.2)$
  \item Cutpoint differences: $c_{j} - c_{j-1} \sim \text{LogNormal}(-0.5, 0.3)$ for $j = 2, \dots, K-1$
\end{itemize}

To ensure ordered and well-separated cutpoints, the cutpoint differences are shifted by a small constant before summing: $\Delta_j = (c_j - c_{j-1}) + 0.3$.

\newpage
\section{Supplementary figures: Model comparisons}
\label{sec:model-comparison}

\begin{figure}[htbp]
    \centering
    \includegraphics[width=0.9\textwidth]{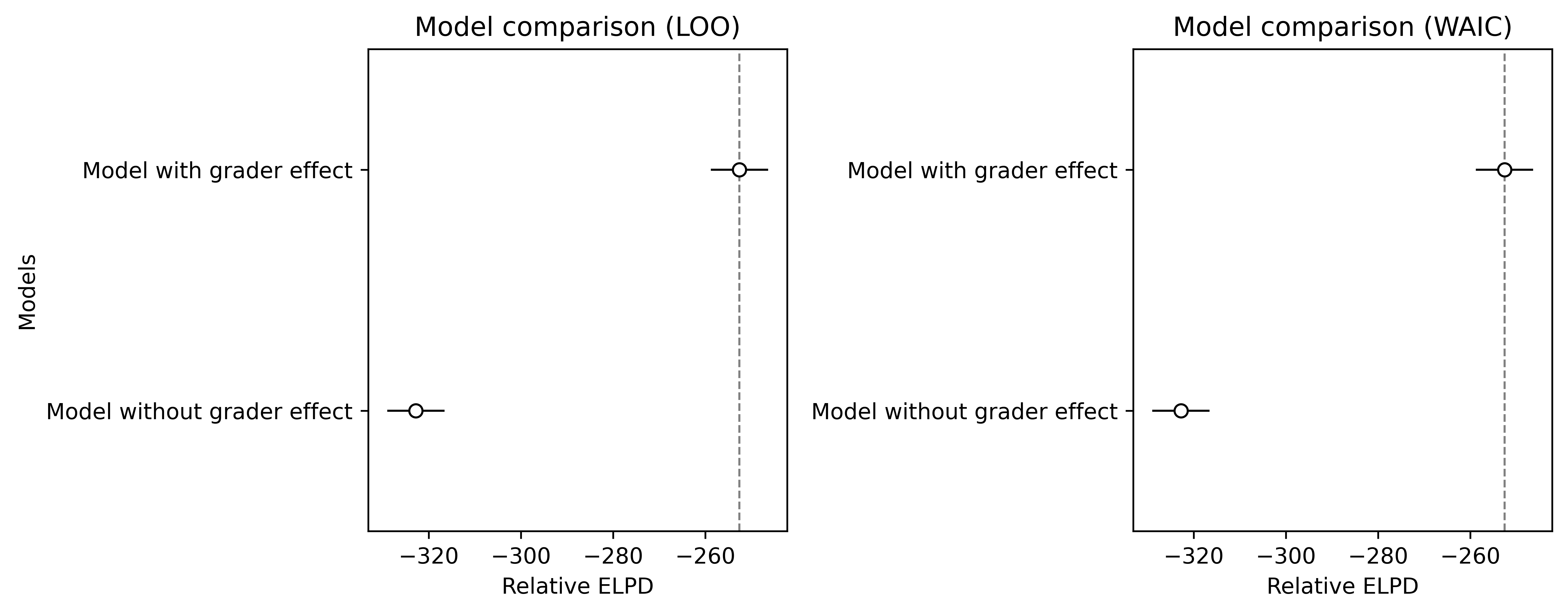}
    \caption{Model comparison for the statistical analysis in \hyperref[sec:q11]{Question 1.1} (quantifying the mean difference between scores assigned by an autograder and a human grader). Left panel: Leave-One-Out cross-validation (LOO) scores. Right panel: Widely Applicable Information Criterion (WAIC). Both metrics approximate the Expected Log Predictive Density (ELPD), a measure of predictive accuracy (higher values indicate better performance). Comparing these models helps determine whether including a grader-effect term (\cref{eq:model_autograder}) is justified by improved predictive performance. In this case, the model with the grader effect clearly outperforms the null model, supporting closer examination of the grader’s impact on scoring}.
    \label{fig:model_selection_Q1}
\end{figure}

\begin{figure}[htbp]
    \centering
    \includegraphics[width=0.9\textwidth]{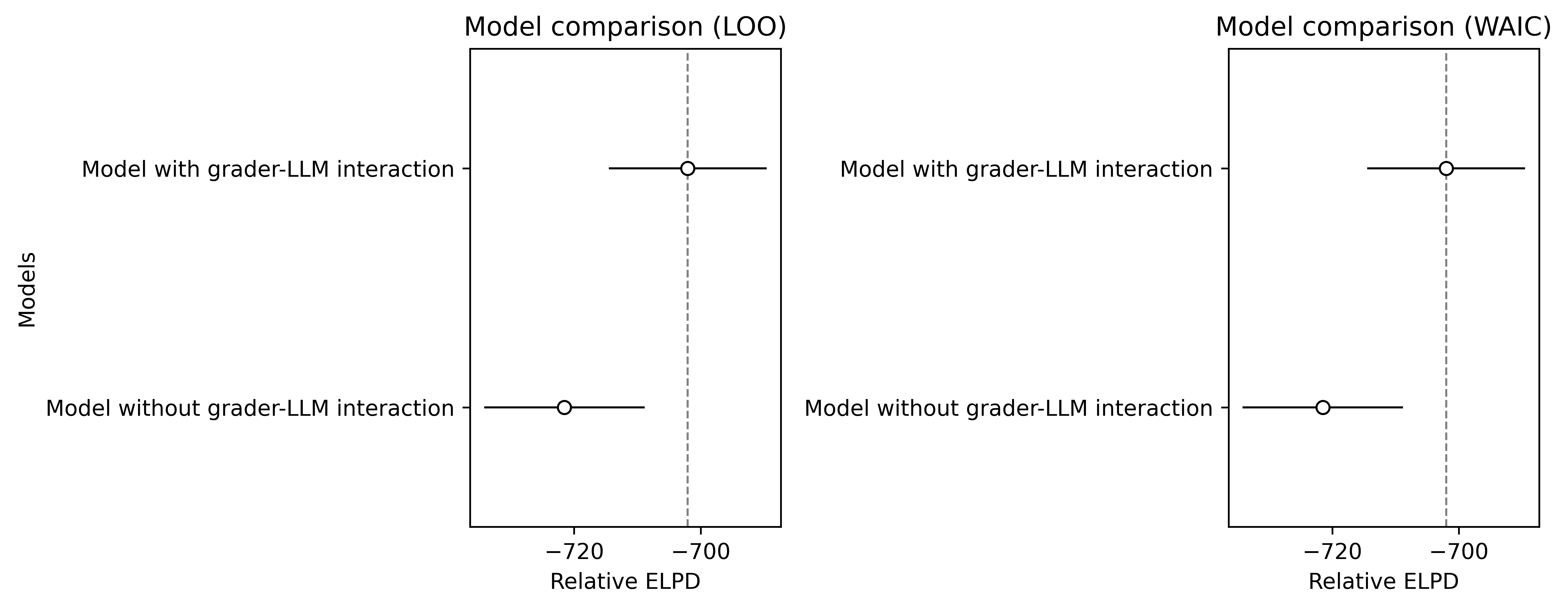}
    \caption{Model comparison for the statistical analysis in \hyperref[sec:q2]{Question 2} (Do autograders favour their own generation?). Left panel: Leave-One-Out cross-validation (LOO) scores. Right panel: Widely Applicable Information Criterion (WAIC). Both metrics approximate the Expected Log Predictive Density (ELPD), a measure of predictive accuracy (higher values indicate better performance). Comparing models helps determine whether the added complexity of including an interaction term is justified by improved predictive performance. In this case, the model with the grader–LLM interaction (\cref{eq:model_autograder_llm_interaction}) performs slightly better than the model without interaction, supporting a closer examination of potential self-bias effects.}
    \label{fig:model_selection_Q2}
\end{figure}

\begin{figure}[htbp]
    \centering
    \includegraphics[width=0.9\textwidth]{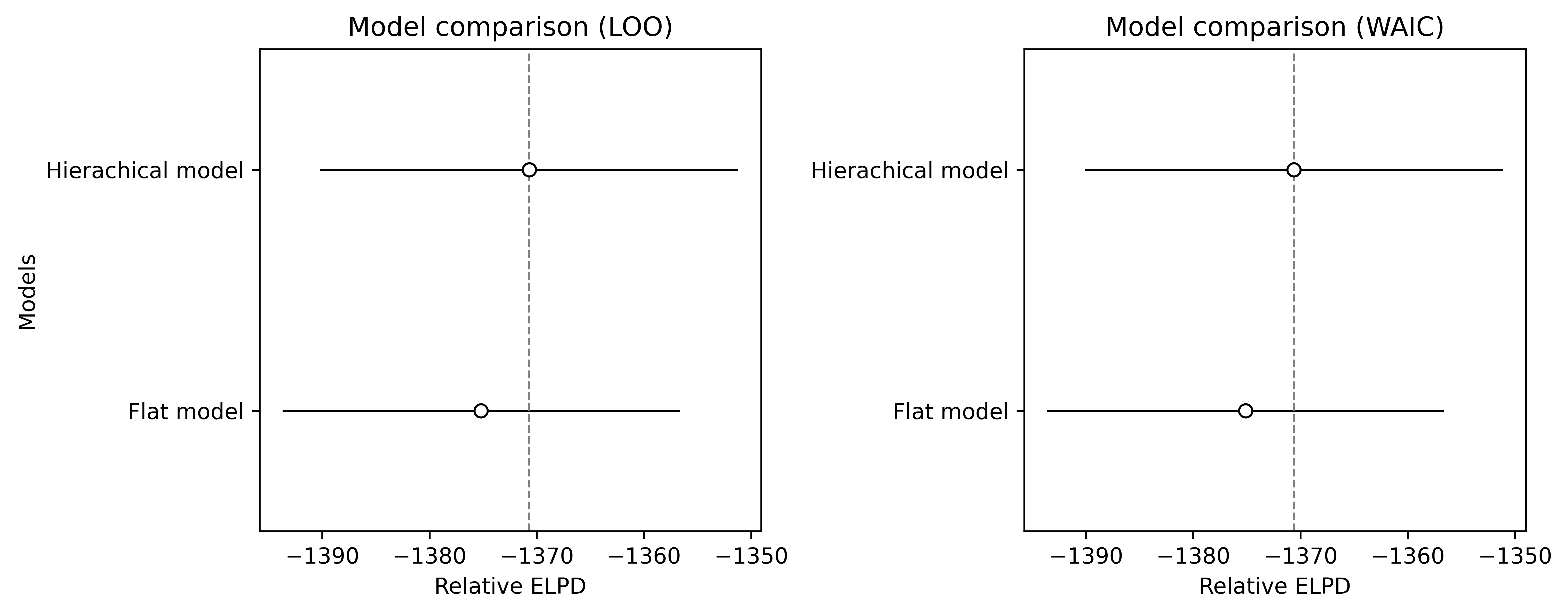}
    \caption{Model comparison for the statistical analysis in \hyperref[sec:q3]{Question 3} (Do autograders differ systematically from human experts?). Left panel: Leave-One-Out cross-validation (LOO) scores. Right panel: Widely Applicable Information Criterion (WAIC). Both metrics approximate the Expected Log Predictive Density (ELPD), a measure of predictive accuracy (higher values indicate better performance). The models perform similarly, which is expected given that the data is simulated without an explicitly hierarchical structure. Here we choose the hierarchical model (\cref{eq:model_hierarchical}) to demonstrate how to interpret its parameters. In practice, when models perform similarly, researchers should favour the simpler model unless theoretical or interpretability considerations justify the added complexity.}
    \label{fig:model_selection_Q3}
\end{figure}

\begin{figure}[htbp] 
    \centering 
    \includegraphics[width=0.9\textwidth]{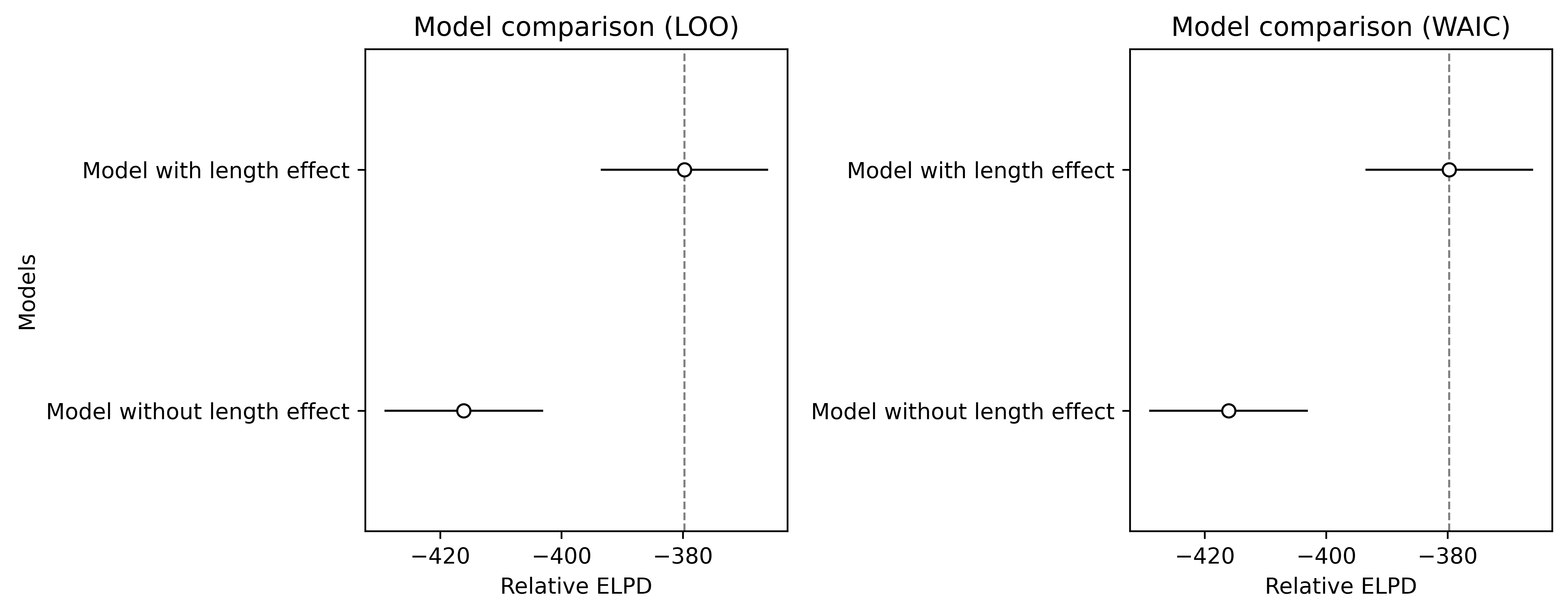} \caption{Model comparison for the statistical analysis in \hyperref[sec:q5]{Question 5} (Do autograders favour longer outputs?). Left panel: Leave-One-Out cross-validation (LOO) scores. Right panel: Widely Applicable Information Criterion (WAIC). Both metrics approximate the Expected Log Predictive Density (ELPD), a measure of predictive accuracy (higher values indicate better performance). Comparing these models tests whether including a length-effect term (\cref{eq:model_pairwise_hierarchical}) significantly improves predictive performance. The model including the length effect clearly outperforms the simpler model without it, justifying a closer investigation into grader-specific length biases.} 
    \label{fig:model_selection_Q5} 
\end{figure}

\newpage
\section{Supplementary question: Is the grading scale well calibrated?}
\label{sec:grading_scale}

In this section we explore the grading process, and why we chose an ordered logistic regression model. To ensure consistent evaluation, Florence needs to establish standardised scoring across all graders. One simple approach would be to instruct everyone (humans and autograders) to count the number of relevant keywords from a predefined list. Besides obvious issues (e.g., synonyms need to be accounted for), this approach is not scalable as it require building explicit keyword lists for each open-ended question. Florence, as often done in practice, will instead develop a grading rubric that can be applied to each question. She might come up with something like in \cref{tab:rubric}.

\begin{table}[h]
\centering
\renewcommand{\arraystretch}{1.2}
\begin{tabular}{c|>{\raggedright\arraybackslash}p{11cm}}
\toprule
\textbf{Points} & \textbf{Description} \\
\midrule
1  & Completely off-topic or no relevant content. \\
2  & Minimal response with no clear concepts or severe confusion. \\
3  & Mentions a relevant idea but largely undeveloped or inaccurate. \\
4  & States one relevant concept with limited clarity or major misconceptions. \\
5  & Mentions key concepts but lacks depth and contains notable flaws. \\
6  & Covers some key concepts with partial accuracy and development. \\
7  & Addresses key concepts clearly, with minor omissions or inaccuracies. \\
8  & Explains most important concepts with accuracy and reasonable depth. \\
9  & Thorough and accurate response with clear development and insight. \\
10 & Complete responses with deep understanding and insightful connections. \\
\bottomrule
\end{tabular}
\vspace{0.7em} 
\caption{Example of a rubric score that a researcher might create to grade open-ended question.}
\label{tab:rubric}
\end{table}

Having an ordinal scale, instead of just a description, is of course very useful. But this scale is not as simple to interpret as an interval scale (e.g., temperature). If Florence wants to make claims like ``autograder A increases the score by 1 point,'' she needs to be aware of whether the intervals between categories are equivalent. If they are not (which is often the case with ordinal scales), a 1-point increase will mean something different depending on where it occurs on the scale. For example, moving from 5 to 6 might represent crossing some fundamental threshold, while moving from 9 to 10 might just be a small qualitative improvement between two already good responses. 

In an ordered logistic regression model, the scores are considered individual categories with a meaningful order. The model maps the observed scores onto a continuous latent scale and, if necessary, can estimate the category boundary values (called cutpoints) on this latent scale. By examining the spacing between these cutpoints, we can determine whether the intervals in our grading scale are equivalent across the range. If they are not, we can make inferences from the distances between cutpoints. For example, if some cutpoints are widely spread, it could indicate that we are capturing an important threshold (i.e., substantial changes in the underlying latent score is required to move between categories), that there is a gap in the measurement scale or that graders are reluctant to use certain portions of the scale. Conversely, if some cutpoints are close together, it could indicate that the scale is very sensitive in that region (i.e., small changes in the latent score result in different observed scores), that models have similar latent abilities in that ``region'' or that the scale contains redundant categories. 

Building great scales is by no means an easy task. It is a well-established challenge that has been thoroughly studied in the field of psychometrics, but is beyond the scope of the presented paper. However, using a GLM with an ordered logistic regression is a useful tool to examine the properties of a given scale and can help identify areas that require caution during interpretation. 

Going back to Florence, to better capture differences in model capabilities, she decides to examine her grading scale. To do this, she can look at the learned cutpoint parameters of the ordered logistic regression. Taking \cref{eq:model_autograder_llm} for example, we can write the cumulative probability more explicitly as:   

\begin{equation}
\begin{gathered}
\phi_i = \beta_0 + \beta_1 \cdot X_i^{\text{grader}} + \beta_2 \cdot X_i^{\text{LLM}} \\
p_{ij} = \text{logit}^{-1}(c_j - \phi_i)
\end{gathered}
\label{eq:model_cutpoint_inspection}
\end{equation}

where $\phi_i$ is the linear predictor, representing the location of response $i$ on an unobserved latent scale. This latent scale is assumed to underlie the observed ordinal scores (e.g., 1–10). The cutpoints $c_j$ divide the latent scale into discrete intervals corresponding to the observed score categories. The probability $p_{ij}$ represents the cumulative probability that the score assigned to response $i$ is less than or equal to category $j$, and is calculated as the inverse logit of the difference between the cutpoint $c_j$ and the linear predictor $\phi_i$.

Conceptually, this means that the probability of scoring at (or below) a certain grade j (on the observed scale) is calculated by comparing the linear predictor to the cutpoint (on the latent scale). These cutpoints are the mechanism through which ordered logistic models maintain the ordinal structure of the data. They were therefore naturally present in the previously discussed models, but were not mentioned as the focus was on the predictor variables. 

Once Florence fits this model, she can analyse the inferred cutpoint values. Those can be found in \cref{tab:cutpoints}. She observes that the distance c1-c2, and c2-c3, is below 1 unit, whereas the distance c6-c7 and c7-c8 is above 1.5 units. This large jump suggests that her scale lacks sensitivity around c6-c8 (i.e., many performance scores are clustering there). From this, she could decide that she wants to better capture difference in that region and therefore adds some intermediate categories in this area.

\begin{table}[h!]
\centering
\renewcommand{\arraystretch}{1.2}
\begin{tabular}{
    >{\centering\arraybackslash}p{1.3cm}|
    r|
    r|
    c|
    >{\raggedright\arraybackslash}p{2cm}|
    >{\raggedright\arraybackslash}p{5cm}
}
\toprule
\textbf{Cutpoint} & \textbf{Latent value} & \makecell{\textbf{Latent} \\ \textbf{interval size}} & \textbf{Score range} & \makecell{\textbf{Score range} \\ \textbf{threshold}} & \textbf{Interpretation} \\
\midrule
c1  & -4.07 & --    & 0--1   & Starting point & Minimum threshold to achieve score of 1 \\
c2  & -3.25 & 0.82  & 1--2   & Narrow         & Small improvement needed to move from 1 to 2 \\
c3  & -2.39 & 0.86  & 2--3   & Narrow         & Small improvement needed to move from 2 to 3 \\
c4  & -1.28 & 1.11  & 3--4   & Moderate       & Moderate improvement needed to move from 3 to 4 \\
c5  & -0.20 & 1.08  & 4--5   & Moderate       & Moderate improvement needed to move from 4 to 5 \\
c6  & 1.29  & 1.49  & 5--6   & Wide           & Large improvement needed to move from 5 to 6 \\
c7  & 3.11  & 1.82  & 6--7   & Wide           & Large improvement needed to move from 6 to 7 \\
c8  & 4.71  & 1.60  & 7--8   & Wide           & Large improvement needed to move from 7 to 8 \\
c9  & 5.60  & 0.89  & 8--9   & Narrow         & Small improvement needed to move from 8 to 9 \\
c10 & 6.22  & 0.62  & 9--10  & Narrow         & Small improvement needed to move from 9 to 10 \\
\bottomrule
\end{tabular}
\vspace{0.7em} 
\caption{Cutpoints and interpretations for score intervals}
\label{tab:cutpoints}
\end{table}

\end{document}